\documentclass[nohyperref]{article}

\usepackage{microtype}
\usepackage{graphicx}
\usepackage{subfigure}
\usepackage{booktabs} 

\usepackage{hyperref}

\usepackage[accepted]{icml2022}

\usepackage{amsmath}
\usepackage{amssymb}
\usepackage{mathtools}
\usepackage{amsthm}

\usepackage[capitalize,noabbrev]{cleveref}

\theoremstyle{plain}
\newtheorem{theorem}{Theorem}[section]

\theoremstyle{definition}
\newtheorem{definition}[theorem]{Definition}

\theoremstyle{remark}

\usepackage[textsize=tiny]{todonotes}

\newcommand\md{r}
\newcommand\mdr{s}
\newcommand\zgk{z^{1:K}}
\newcommand\rhogk{\rho^{1:K}}

\newcommand\zg{z} 
\newcommand\yg{y}

\newcommand{\E}{\operatornamewithlimits{\mathbb{E}}}
\newcommand{\ug}{\theta}

\newcommand{\EVI}{\mathcal{L}_\mathrm{VI}}
\newcommand{\EUHA}{\mathcal{L}_\mathrm{UHA}}
\newcommand{\EAIS}{\mathcal{L}_\mathrm{AIS}}
\newcommand{\EIW}{\mathcal{L}_\mathrm{IW}}
\newcommand{\EG}{\mathcal{L}}

\usepackage{xspace}

\newenvironment{talign*}
 {\csname align*\endcsname}
 {\endalign}

\newenvironment{tequation}
 {\equation}
 {\endequation}

\definecolor{ao(english)}{rgb}{0.0, 0.5, 0.0}

\icmltitlerunning{Locally Enhanced Bounds for Hierarchical Models}

\begin{document}

\twocolumn[
\icmltitle{Variational Inference with Locally Enhanced Bounds for Hierarchical Models}



\icmlsetsymbol{equal}{*}

\begin{icmlauthorlist}
\icmlauthor{Tomas Geffner}{umass}
\icmlauthor{Justin Domke}{umass}
\end{icmlauthorlist}

\icmlaffiliation{umass}{College of Information and Computer Sciences, University of Massachusetts Amherst, MA, USA.}

\icmlcorrespondingauthor{Tomas Geffner}{tgeffner@cs.umass.edu}
\icmlcorrespondingauthor{Justin Domke}{domke@cs.umass.edu}

\icmlkeywords{variational inference, hierarchical models, annealing, hamiltonian}

\vskip 0.3in
]



\printAffiliationsAndNotice{}  

\begin{abstract}
Hierarchical models represent a challenging setting for inference algorithms. MCMC methods struggle to scale to large models with many local variables and observations, and variational inference (VI) may fail to provide accurate approximations due to the use of simple variational families. Some variational methods (e.g. importance weighted VI) integrate Monte Carlo methods to give better accuracy, but these tend to be unsuitable for hierarchical models, as they do not allow for subsampling and their performance tends to degrade for high dimensional models. We propose a new family of variational bounds for hierarchical models, based on the application of tightening methods (e.g. importance weighting) separately for each group of local random variables. We show that our approach naturally allows the use of subsampling to get unbiased gradients, and that it fully leverages the power of methods that build tighter lower bounds by applying them independently in lower dimensional spaces, leading to better results and more accurate posterior approximations than relevant baselines.
\end{abstract}

\section{Introduction}

\begin{figure*}[t]
\centering
\input{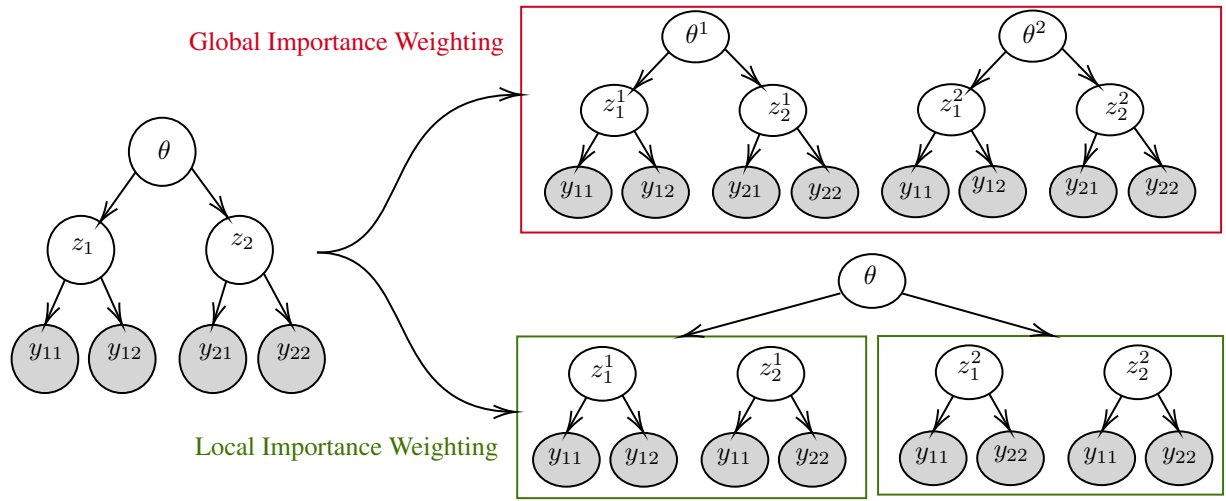}
\caption{Tighter bounds using importance weighting. \textcolor{red}{A direct (global) application of importance weighting} generates independent set of copies of all variables in the model to build a tighter bound. \textcolor{ao(english)}{A local application of importance weighting}, on the other hand, generates copies at the local level and applies importance weighting separately for each group of local variables to build the \textit{locally-enhanced bound}. Gray nodes represent observed variables, which are fixed  (not re-sampled with every generated copy).}
\label{fig:intuition_IW}
\end{figure*}

Hierarchical models \cite{kreft1998introducing, gelman2006multilevel, snijders2011multilevel} represent a general class of probabilistic models which are used in a wide range of scenarios. They have been successfully applied in psychology \cite{vallerand1997toward}, ecology \cite{royle2008hierarchical, cressie2009accounting}, political science \cite{lax2012democratic}, collaborative filtering \cite{lim2007variational}, and topic modeling \cite{blei2003latent}, among others. While these models may take a wide range of forms, a widely used one consists of a tree structure, where a set of global variables $\ug$ controls the distribution over local variables $z_i$ in multiple independent groups (see \cref{fig:intuition_IW} (left)). Then, after observing some data $y_i$ from each group, the inference problem consists in accurately approximating the posterior distribution over the global and local variables.

Inference is often difficult in hierarchical models. MCMC methods struggle to scale to big models and datasets due to their incapacity to handle subsampling \cite{betancourt2015fundamental, bardenet2017markov}. Variational inference (VI) methods, on the other hand, are naturally compatible with subsampling, and thus represent a more scalable alternative \cite{hoffman2013stochastic, doublystochastic_titsias, agrawal2021amortized}. Their accuracy, however, is sometimes limited by the use of simple variational families, such as factorized Gaussians.

Recently, many methods have been proposed to integrate Monte Carlo methods into variational inference to give tighter bounds and better posterior approximations (henceforth \textit{tightening methods}). These include importance weighting \cite{IWVAE} and many others \cite{salimans2015markov, wolf2016variational, maddison2017filtering, domke2019divide, thin2021monte, dais, uha}. While these methods have shown good performance in practice, we observe that a direct application of them may be unsuitable with hierarchical models. There are two reasons for this. First, the posterior distributions for hierarchical models are often high dimensional---the dimensionality typically grows linearly with the number of local variables. This is problematic as the performance of tightening methods sometimes degrades in higher dimensions. Second, current tightening methods are incompatible with subsampling, leading to slow inference.


Practitioners are thus faced with a choice: They can use powerful but inefficient methods (variational inference with tightening methods), or faster methods with lower accuracy (plain variational inference).

We propose \textit{locally-enhanced bounds}, a new family of variational objectives for hierarchical models that enjoys much of the best of both worlds. The main idea involves applying tightening methods at a local level, separately for each set of local variables $z_i$, while using a regular variational approximation (e.g. Gaussian, normalizing flow) to model the posterior distribution over the global variables. This is naturally compatible with subsampling, making inference more efficient. Additionally, it maintains much of the benefit of tightening methods in terms of improved bounds and more accurate posterior approximations \cite{domke2019divide}. We show the intuition behind our method in \cref{fig:intuition_IW}. 


We present an extensive empirical evaluation of our approach using two tightening methods: importance weighting \cite{IWVAE} and uncorrected Hamiltonian annealing \cite{uha, dais}. The former is based on importance sampling, while the latter uses Hamiltonian Monte Carlo \cite{neal2011mcmc, betancourt2017conceptual} transition kernels to build an enhanced variational distribution. We observe empirically that the proposed approach yields better results than plain variational inference and a traditional application of tightening methods. 


\section{Preliminaries}

\subsection{Hierarchical Models} \label{sec:hm}


While hierarchical models may take a wide range of forms \cite{gelman2006data}, in this work we focus on a two-level formulation, using $\ug$ to denote the global variables, and $z_i$ and $y_i$ to denote the local variables and observations of group $i$. By letting $\zg = (z_1, \hdots, z_M)$ and $\yg = (y_1, \hdots, y_M)$, the corresponding probabilistic model is given by
\begin{align}
    p(\ug, \zg, \yg) & = p(\ug) \prod_{i=1}^M p(z_i, y_i \vert \ug), \label{eq:hm_simpleobs}
\end{align}
where the exact form of $p(z_i,y_i\vert \ug)$ depends on the application. Often, $y_i$ is conditionally independent of $\ug$ given $z_i$, and so $p(z_i, y_i \vert \ug) = p(y_i \vert z_i) p(z_i \vert \ug)$. In addition, $y_i$ often consists of $N_i$ observations $y_{i1}, \hdots, y_{iN_i}$ that are conditionally independent given $z_i$, and so $p(y_i|z_i) = \prod_{j=1}^{N_i} p(y_{ij}\vert z_i)$. However neither of these simplifications is required. 


\subsection{Variational Inference}

Variational Inference is a popular method used to approximate posterior distributions. Given some model $p(z, y)$, where $y$ is observed data and $z$ latent variables, the goal of variational inference is to find a simpler distribution $q(z)$ to approximate the target $p(z|y)$ \cite{jordan1999introduction, wainwright2008graphical, blei2017variational}. VI does this by finding the parameters of $q(z)$ that maximize the evidence lower bound (ELBO), a lower bound on the log-marginal likelihood $\log p(y)$, given by
\begin{equation}
\EVI(q(z) \Vert p(z, y)) = \E_{q(z)} \log \frac{p(z, y)}{q(z)}. \label{eq:plainVIbound}
\end{equation}
It can be shown that this is equivalent to minimizing the KL-divergence from $q(z)$ to the true posterior $p(z\vert y)$.

\subsection{Tighter Bounds for Variational Inference} \label{sec:ptb}

While VI has been successfully applied in a wide range of tasks \cite{blei2017variational}, its performance its sometimes limited by the use of simple approximating families for $q(z)$, such as Gaussians. A popular approach to address this drawback involves using tighter lower bounds on the log-marginal likelihood \cite{IWVAE, dais}, which lead to better posterior approximations \cite{domke2018importance, domke2019divide}.

\textbf{Importance Weighting (IW)} \cite{IWVAE, cremer2017reinterpreting} uses $K$ samples $z^k \sim q(z^k)$ to build a lower bound on the log-marginal likelihood $\log p(y)$ as
\begin{equation}
\EIW^K(q(z) \Vert p(z, y)) = \E_{\prod_k q(z^k)} \log \frac{1}{K} \sum_{k=1}^K \frac{p(z^k, y)}{q(z^k)}, \label{eq:IWboundVI}
\end{equation}
which is provably tighter that the variational inference bound from \cref{eq:plainVIbound} for any $K > 1$.

\textbf{Annealed Importance Sampling (AIS)} \cite{neal2001ais} is another method that can be used to build tighter bounds. It defines a sequence of $K-1$ (unnormalized) densities $\pi^1(z), \hdots, \pi^{K-1}(z)$ that gradually bridge from $q(z)$ to $p(z, y)$. Then, it augments $q(z)$ using MCMC transitions $T^k(z^{k+1}\vert z^k)$ that hold the corresponding bridging density $\pi^k$ invariant, and builds a lower bound on $\log p(y)$ as\footnote{For simplicity, the notation for the transitions and bridging densities ignores their dependency on $y$, which is fixed.}
\begin{multline}
\EAIS^K(q(z)\Vert p(z, y)) = \\ \E_{q(\zgk)} \log \frac{p(z^K, y)}{q(z^1)} \prod_{k=1}^{K-1} \frac{\pi^k(z^k)}{\pi^{k+1}(z^{k+1})},
\end{multline}
where $z^k$ represents the $k$-th variable generated by the MCMC-augmented sampling process. It has been observed that AIS with Hamiltonian Monte Carlo kernels \cite{neal2011mcmc, betancourt2017conceptual} often yields tight lower bounds in practice \cite{sohl2012hamiltonian, grosse2015sandwiching, wu2016quantitative}. However, since the HMC transitions include a correction step, the resulting bound is not differentiable. Thus, low variance reparameterization gradients cannot be used to tune the method's many parameters.

\textbf{Uncorrected Hamiltonian Annealing (UHA)} \cite{uha, dais} is a method that addresses the non-differentiability drawback suffered by Hamiltonian AIS. It does so by mimicking the construction used by Hamiltonian AIS, but using uncorrected HMC kernels for the transitions. Then, UHA builds a differentiable lower bound on the log-marginal likelihood $\log p(y)$ as
\begin{multline}
\EUHA^K(q(z)\Vert p(z, y)) = \\ \E_{q(\zgk, \rhogk)} \log \frac{p(z^K, y)}{q(z^1)} \prod_{k=1}^{K-1} \frac{\md(\rho^{k+1})}{\md(\tilde \rho^k)}, \label{eq:bounduhaVI}
\end{multline}
where $\rho^k$ and $\tilde \rho^k$ are the momentum variables generated by HMC at each step $k$, and $z^1$ and $z^K$ are the first and last samples from the chain, respectively. We give full details on AIS and UHA in \cref{app:boundsdets}.

All these methods have been observed to provide lower bounds that are significantly tighter than the original one used by variational inference, resulting in better posterior approximations \cite{domke2019divide}.

\section{Locally Enhanced Bounds for Hierarchical Models} \label{sec:method}

This section introduces our method. We begin with a brief description on the use of variational inference and tightening methods for hierarchical models and their limitations. We then introduce our new family of variational bounds, \textit{locally-enhanced bounds}, which addresses these limitations.

\subsection{Variational Inference for Hierarchical Models} \label{sec:VIHM}

Given a hierarchical model $p(\ug, \zg, \yg)$ and some observations for $\yg$, we can approximate the posterior distribution $p(\ug, \zg \vert \yg)$ using VI with a variational distribution $q(\ug, \zg)$. There are many potential choices for the approximating distribution. One could use, for instance, a Gaussian with a diagonal or dense covariance \cite{challis2013gaussian}. However, best results have been observed using a distribution that follows the true posterior's factorization \cite{hoffman2015stochastic, agrawal2021amortized}
\begin{align} \label{eq:poststruct}
q(\ug, \zg) = q(\ug) \prod_{i=1}^M q(z_i\vert \ug),
\end{align}
which explicitly avoids modeling dependencies not present in the target posterior. Then, the parameters of $q(\ug, \zg)$ are trained by maximizing the objective
\begin{multline}
\EVI(q(\ug, \zg) \Vert p(\ug, \zg, \yg)) = \\
\E_{q(\ug, \zg)} \bigg[ \underbrace{\log \frac{p(\ug)}{q(\ug)}}_{\mbox{global term}} + \sum_{i=1}^M \underbrace{\log \frac{p(z_i, y_i\vert \ug)}{q(z_i\vert \ug)}}_{\mbox{local terms}} \bigg]. \label{eq:elbohm}
\end{multline}
While computing this objective's exact gradient is typically intractable, an unbiased estimate can be efficiently obtained by applying the reparameterization trick \cite{vaes_welling, rezende2014stochastic, doublystochastic_titsias} and subsampling $M' < M$ local terms.\footnote{The reparameterization trick is applicable for distributions $q(\ug, \zg)$ parameterized by $w$ for which the sampling process can be divided in two steps: Sampling a $w$-independent noise variable $\epsilon\sim q_0(\epsilon)$, and then obtaining the sample $(\ug, \zg)$ as a $w$-dependent differentiable transformation $(\ug, \zg) = \mathcal{T}_w(\epsilon)$.}
The fact that VI allows for subsampling makes the method a particularly attractive choice for cases where the number of groups $M$ is large.

\subsection{Unsuitability of Tightening Methods for Hierarchical Models} \label{sec:tbHMbad}

One can directly apply tightening methods to variational inference for hierarchical models. However, this may work poorly when the number of groups $M$ is large. This is because some of these methods may provide less tightening in high dimensions (particularly importance weighting \cite{bengtsson2008curse, chatterjee2018sample}) and because they are not compatible with subsampling, which makes inference less efficient. This can be seen, for instance, by considering the importance weighting objective for hierarchical models
\begin{multline}
\EIW^K(q(\ug, \zg\Vert p(\ug, \zg, \yg))) = \\
\E_{\prod_k q(z^k)} \left[ \log \frac{1}{K} \sum_{k=1}^K \frac{p(\ug^k)}{q(\ug^k)} \prod_{i=1}^M \frac{p(z_i^k, y_i\vert \ug^k)}{q(z_i^k\vert \ug^k)} \right]. \label{eq:naiveIW}
\end{multline}
There does not appear to be any way to estimate the objective above subsampling $M'<M$ groups without introducing bias. This is problematic for stochastic optimization, as using no subsampling leads to expensive gradient evaluations and a slow overall optimization process, but the use of biased gradients may lead to suboptimal parameters \cite{MarkovianSC, biasedalphafail} or may even cause the optimization process to diverge \cite{ajalloeian2020convergence}.

Importance weighting is not unique in this regard. Other methods, such as annealed importance sampling or uncorrected Hamiltonian annealing, also lead to objectives that do not allow unbiased subsampling either \cite{dais}.

\subsection{Variational Inference with Locally Enhanced Bounds} \label{sec:locallyenhancedbounds}

This section introduces our method. Our goal is to apply tightening methods to boost variational inference's performance on hierarchical models while avoiding the aforementioned issues. We propose to achieve this by applying tightening methods only for the local variables, separately for each group $i=1, \hdots, M$. This leads to a new family of variational objectives, which we call \textit{locally-enhanced bounds}. Our construction of this new family of bounds is based on the concept of a bounding operator.

\begin{definition}[Bounding operator]
An operator $\EG(\cdot \Vert \cdot)$ is a bounding operator if, for any distributions $q(z)$ and $p(z, y)$, it satisfies $\EG(q(z) \Vert p(z, y)) \leq \log p(y)$. 
\end{definition}

Example bounding operators we have seen so far include plain VI (\cref{eq:plainVIbound}), importance weighted VI (\cref{eq:IWboundVI}), and uncorrected Hamiltonian annealing (\cref{eq:bounduhaVI}).

Building locally-enhanced bounds is simple. We begin by observing that the typical objective used by VI with hierarchical models, shown in \cref{eq:elbohm}, can be re-written as 
\begin{multline}
\EVI(q(\ug, \zg) \Vert p(\ug, \zg, \yg))  \\
= \E_{q(\ug)} \left[ \log \frac{p(\ug)}{q(\ug)} + \sum_{i=1}^M \E_{q(z_i\vert \ug)}\log \frac{p(z_i, y_i\vert \ug)}{q(z_i\vert \ug)} \right]  \\
= \E_{q(\ug)} \left[ \log \frac{p(\ug)}{q(\ug)} + \sum_{i=1}^M \EVI(q(z_i\vert \ug) \Vert p(z_i, y_i\vert \ug))\right]. \label{eq:rewiritingHELBO}
\end{multline}

Then, a locally-enhanced bound is obtained by replacing $\EVI$ in the last line of \cref{eq:rewiritingHELBO} by any other bounding operator. One could use any of the tightening techniques described in \cref{sec:ptb}, such as importance weighting or uncorrected Hamiltonian annealing. (One could also use AIS, though this yields a non-differentiable objective.)
The following theorem shows that bounds constructed this way always yield valid variational objectives.

\begin{theorem} \label{thm:localbound}
Let $p(\ug, \zg, \yg) = p(\ug) \prod_{i=1}^M p(z_i, y_i\vert \ug)$ and $q(\ug, \zg) = q(\ug) \prod_{i=1}^M q(z_i\vert \ug)$ be any distributions, and let $\EG(\cdot \Vert \cdot)$ be a bounding operator. Then,
\begin{equation}
\E_{q(\ug)} \left[ \log \frac{p(\ug)}{q(\ug)} + \sum_{i=1}^M \EG(q(z_i\vert \ug)\Vert p(z_i, y_i \vert \ug)) \right] \leq \log p(\yg). \label{eq:generic_lowerbound}
\end{equation}
In particular, the gap in the above inequality is
\begin{multline}
\mathrm{KL}(q(\ug) \Vert p(\ug \vert y)) + \\ \sum_{i=1}^M \E_{q(\ug)} \left[ \log p(y_i\vert \ug) - \EG(q(z_i\vert \ug) \Vert p(z_i, y_i\vert \ug)) \right].\label{eq:gap}
\end{multline}
\end{theorem}

We include a proof in \cref{sec:proof}. \cref{thm:localbound} states that we can build a valid locally-enhanced variational bound by using \cref{eq:generic_lowerbound} with any valid bounding operator $\EG(\cdot\Vert \cdot)$. Some examples include, for instance, $\EIW^K$ or $\EUHA^K$, corresponding to importance weighting and uncorrected Hamiltonian annealing, introduced in \cref{sec:ptb}. Equivalently, this construction can be seen as applying the corresponding tightening method at a local level, separately for each group, as shown in \cref{fig:intuition_IW}.

For concreteness, consider importance weighting and its corresponding bounding operator $\EIW^K$. Following the construction described above, we get a locally-enhanced bound of
\begin{equation}
\E_{q(\ug)} \left[ \log \frac{p(\ug)}{q(\ug)} + \sum_{i=1}^M \E_{q(z_i^{1:K}\vert \ug)} \left[ \log \frac{1}{K} \sum_{k=1}^K \frac{p(z_i^k, y_i \vert \ug)}{q(z_i\vert \ug)}\right]\right]. \label{eq:localIW}
\end{equation}

\paragraph{Benefits of locally-enhanced bounds} The benefits of using locally-enhanced bounds are twofold. First, they naturally allow the use of subsampling. This can be seen by noting that the generic locally-enhanced bound from \cref{eq:generic_lowerbound} can be estimated without bias using a subset of local variables $I \subset \{1, \hdots, M\}$ as
\begin{equation}
\E_{q(\ug)} \left[ \log \frac{p(\ug)}{q(\ug)} + \frac{M}{\vert I \vert}\sum_{i\in I} \EG(q(z_i\vert \ug) \Vert p(z_i, y_i\vert \ug))\right],
\end{equation}
where $\vert I\vert$ denotes the size of the set $I$. This is in contrast to direct applications of tightening methods, which are incompatible with subsampling.

Second, they fully leverage the power of tightening methods by applying them separately for each set of local variables, which are often low-dimensional. In fact, for hierarchical models with $M$ groups, the local variables' dimensionality is, on average, $M$ times smaller than that of the full model. Then, one may expect tightening methods to perform well when applied this way. We verify this empirically in \cref{sec:exps}, where we observe that locally-enhanced bounds obtained with importance weighting tend to be considerably better than those obtained by applying importance weighting directly for the full model all at once.

\paragraph{Tightness of locally-enhanced bounds} \cref{eq:generic_lowerbound} shows that better tightening methods lead to tighter locally-enhanced bounds. However, \cref{eq:gap} states that, even in the ideal case where the perfect bounding operator is available,\footnote{That is $\EG(q(z_i) \Vert p(z_i, y_i)) = \log p(y_i)$.} locally-enhanced bounds are only able to fully close the variational gap if the variational approximation over global variables $q(\ug)$ is a perfect approximation of the true posterior $p(\ug\vert \yg)$. If this is not the case, the application of a perfect tightening method yields a variational gap of $\mathrm{KL}(q(\ug) \Vert p(\ug\vert \yg))$.

Often, this does not represent a significant drawback. In practice, for moderately large datasets, the true posterior over global variables $p(\ug\vert \yg)$ is informed by a large number of observations, and thus we might expect it to concentrate and roughly follow a Gaussian distribution. In such cases, accurate approximations can be obtained using a Gaussian variational family. Moreover, even in cases where the true global posterior is non-Gaussian (e.g. small dataset), one could use a more flexible family for $q(\ug)$, such as normalizing flows \cite{tabak2013family, rezende2015variational}. This can be done efficiently, as the dimensionality of $\ug$ is often moderately low and, more importantly, does not depend on the number of groups $M$ nor number of observations.

\paragraph{Locally-enhanced bounds as divergence minimization} Finally, it is worth mentioning that maximizing a locally-enhanced bound can be equivalently formulated as minimizing a divergence between an augmented target and variational approximation. We give details for this construction in \cref{app:div}.





\begin{figure*}[ht]
  \centering
  \includegraphics[scale=0.4, trim = {0 2.4cm 0 0}, clip]{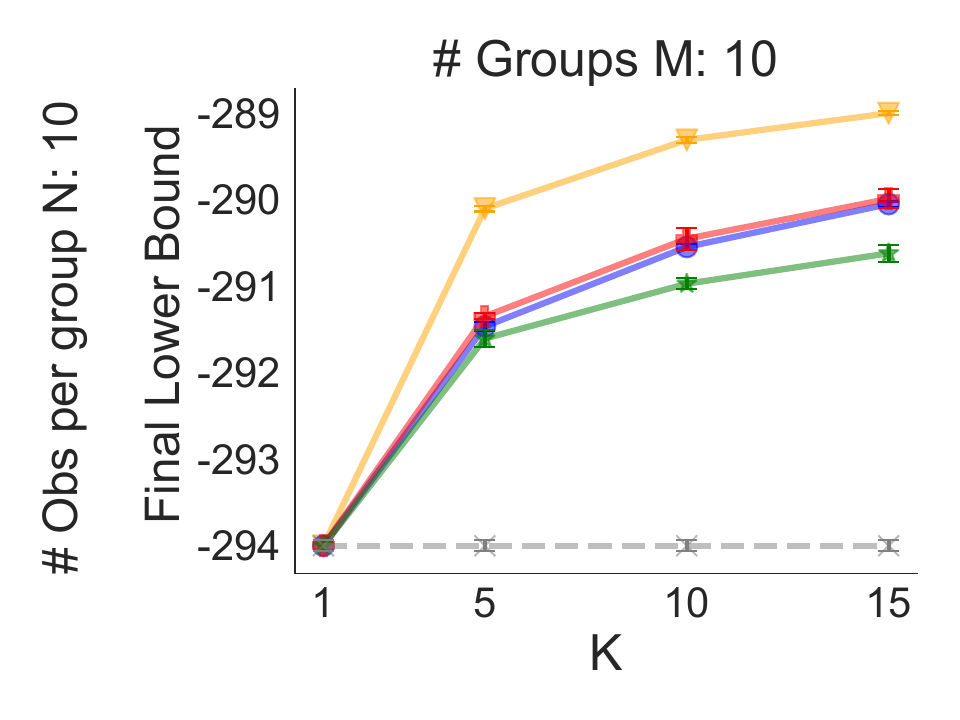}
  \includegraphics[scale=0.4, trim = {3.2cm 2.4cm 0 0}, clip]{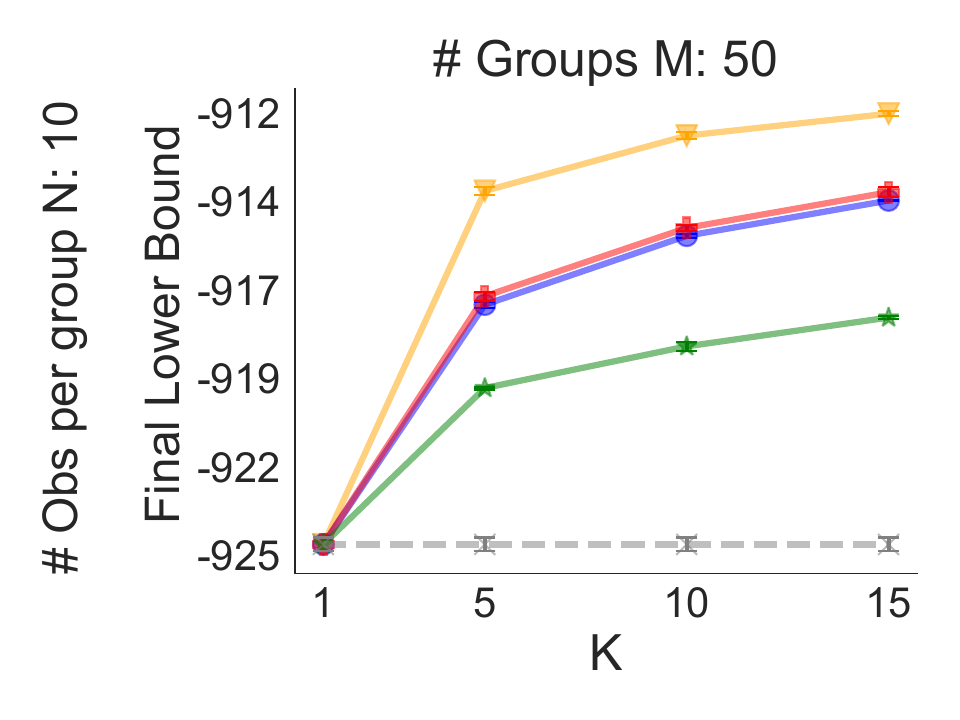}
  \includegraphics[scale=0.4, trim = {3.2cm 2.4cm 0 0}, clip]{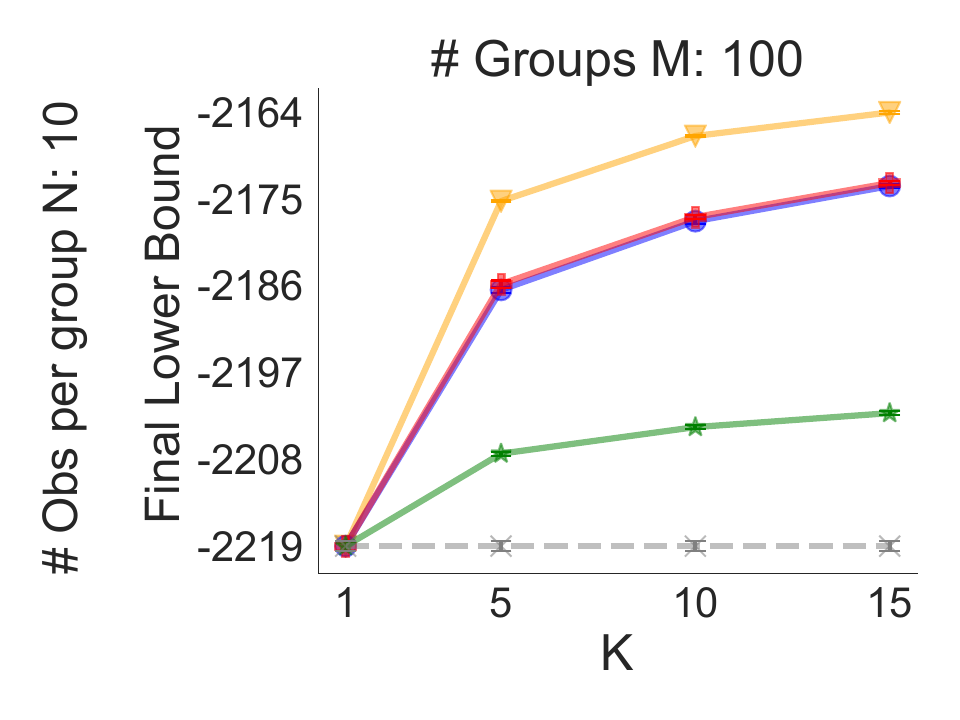}

  \vspace{0.5cm}

  \includegraphics[scale=0.4, trim = {0 0 0 1.5cm}, clip]{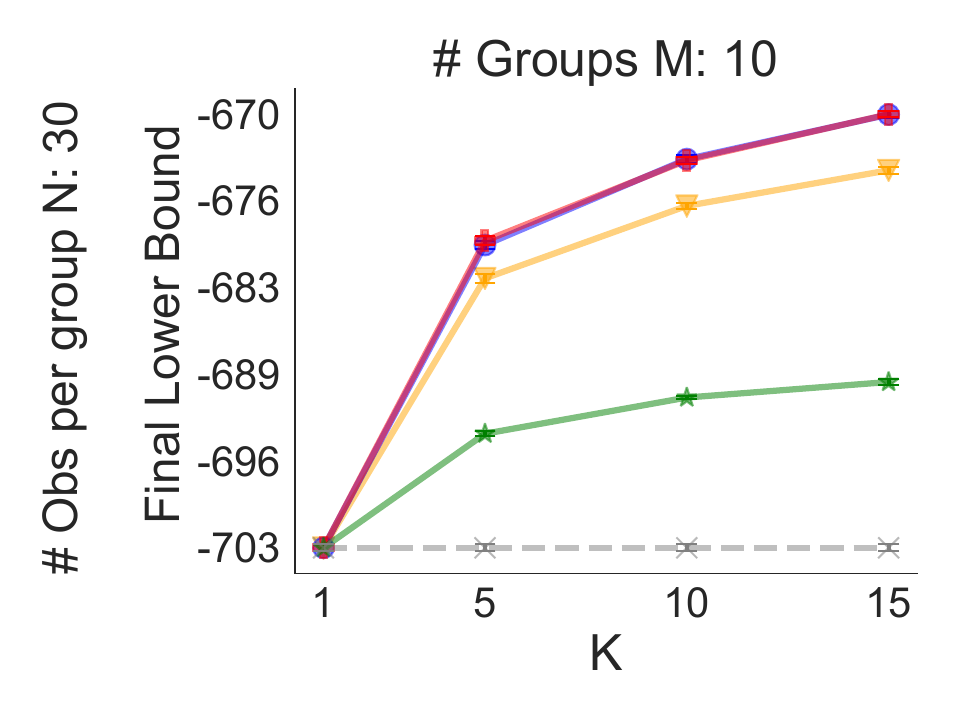}
  \includegraphics[scale=0.4, trim = {3.2cm 0 0 1.5cm}, clip]{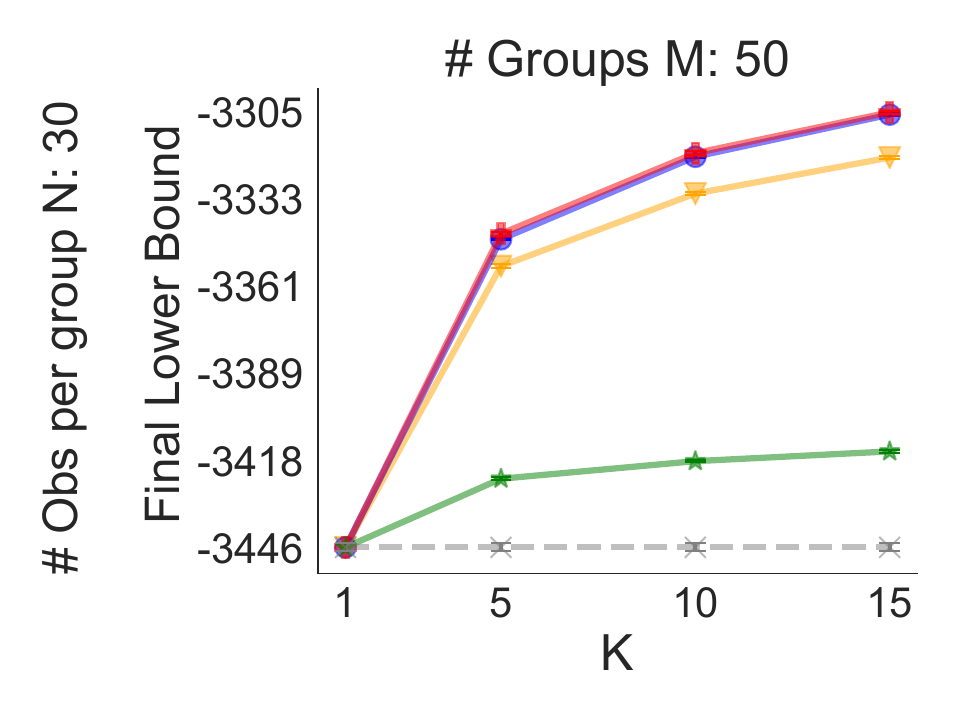}
  \includegraphics[scale=0.4, trim = {3.2cm 0 0 1.5cm}, clip]{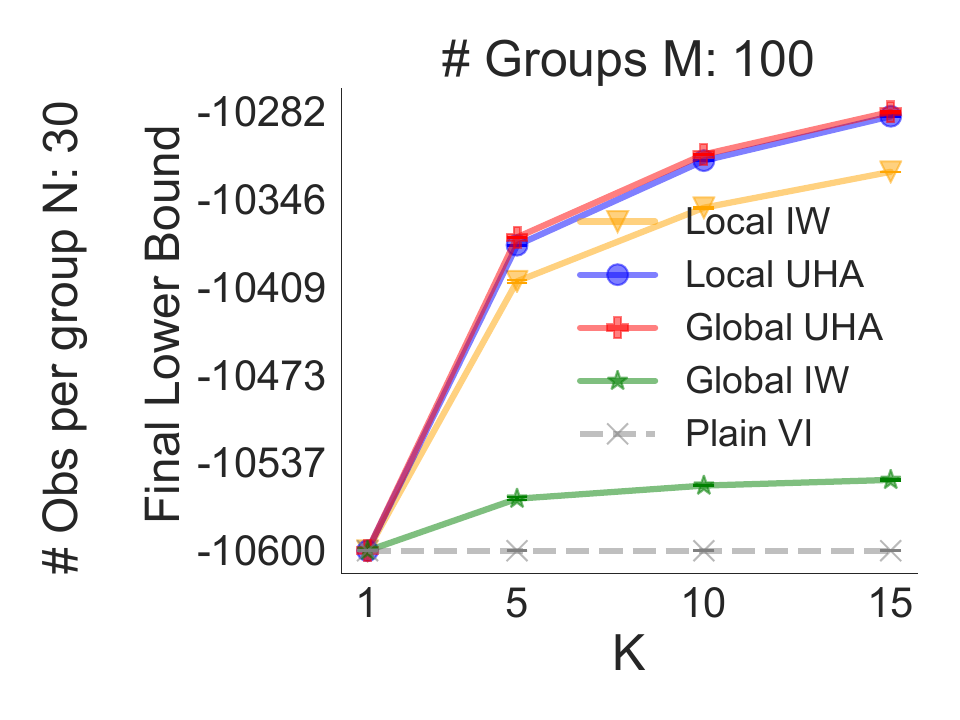}

  \caption{Inference results using locally-enhanced bounds and other baselines on synthetic datasets for varying number of groups (10, 50 and 100) and observations per group (10 and 30). The plots show the final lower bound achieved by different methods after training for 50k steps. All methods (local IW, local UHA, global IW, and global UHA) converge to plain VI for $K=1$. The dimensionality of the local variables $z_i$ is taken to be $d_z=5$ for the datasets with $N=10$ observations per group, and $d_z=20$ for the datasets with $N=30$ observations.}
  \label{fig:results_syn_grid}
\end{figure*}

\section{Related Work}

There is a lot of work on related topics, such as the development of more flexible variational approximations \cite{tabak2013family, rezende2015variational}, better tightening methods \cite{salimans2015markov, domke2019divide}, and more efficient variational methods for hierarchical models \cite{agrawal2021amortized}. Most of these represent contributions orthogonal to ours, and can be used jointly with our locally-enhanced bounds.

Normalizing flows \cite{tabak2013family, rezende2015variational, kingma2016improved, tomczak2016improving}, for instance, are a powerful method to build flexible variational approximations using invertible parametric transformations. Since they typically require a large number of parameters, they are sometimes impractical for very high dimensional problems, such as those that arise when working with hierarchical models with many latent variables. Despite this, they can be used jointly with our method. One could use flows for each local approximation $q(z_i\vert \ug)$ and/or for the global approximation $q(\ug)$, and then optimize their parameters by maximizing a locally-enhanced bound.

Additionally, many powerful tightening methods have been developed \cite{agakov2004auxiliary, IWVAE, salimans2015markov, domke2019divide, uha, dais}. All of these can be easily used to build locally-enhanced bounds following our construction from \cref{sec:locallyenhancedbounds}.

Specifically for hierarchical models, \citet{hoffman2015stochastic} introduced a flexible framework for fast stochastic variational inference. Their approach, however, requires conjugacy, limiting its applicability. To overcome this limitation, \citet{agrawal2021amortized} proposed a parameter efficient algorithm that uses a single amortization network \cite{vaes_welling} to parameterize all local approximations $q(z_i\vert \ug)$. This approach is compatible with our method, as this amortized variational distribution can be used with locally-enhanced bounds.

More generally, \citet{ambrogioni2021automatic} suggested the use of a variational approximation following the prior model's structure. For the hierarchical models considered in this work, this results in a variational approximation with the same structure as the one shown in \cref{eq:poststruct}, which was used by \citet{hoffman2015stochastic, agrawal2021amortized}, and is the one used in this work.

Finally, concurrently with this work, \citet{jankowiak2021surrogate} proposed several promising extensions for uncorrected Hamiltonian Annealing \cite{uha, dais}. One of them involves its use for hierarchical models, applying it independently for each set of local variables. This is equivalent to a locally-enhanced bound built using uncorrected Hamiltonian Annealing. Although this is a minor focus of their work, it can be seen as an instance of our general framework.

\section{Experiments} \label{sec:exps}

  



\begin{figure*}[ht]
  \centering
  \includegraphics[scale=0.4, trim = {0 2.4cm 0 0}, clip]{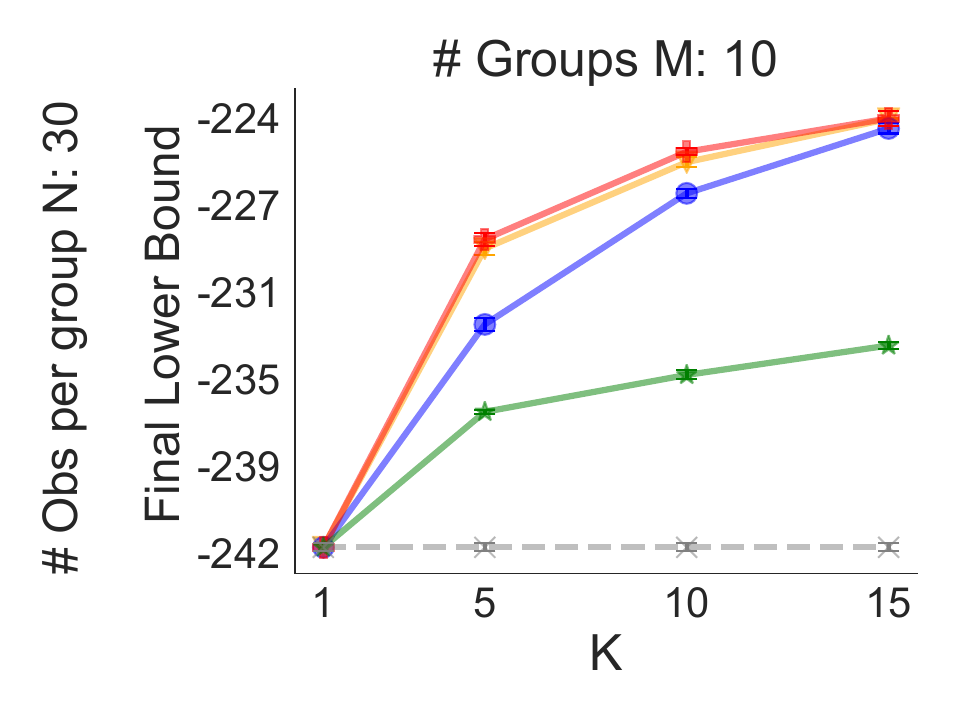}
  \includegraphics[scale=0.4, trim = {3.2cm 2.4cm 0 0}, clip]{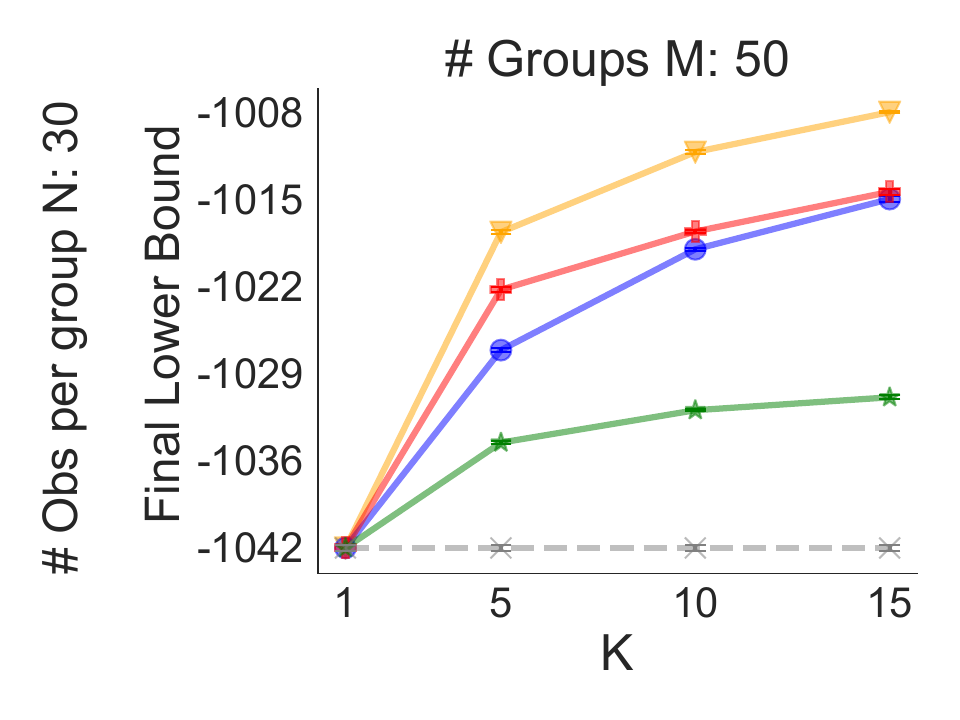}
  \includegraphics[scale=0.4, trim = {3.2cm 2.4cm 0 0}, clip]{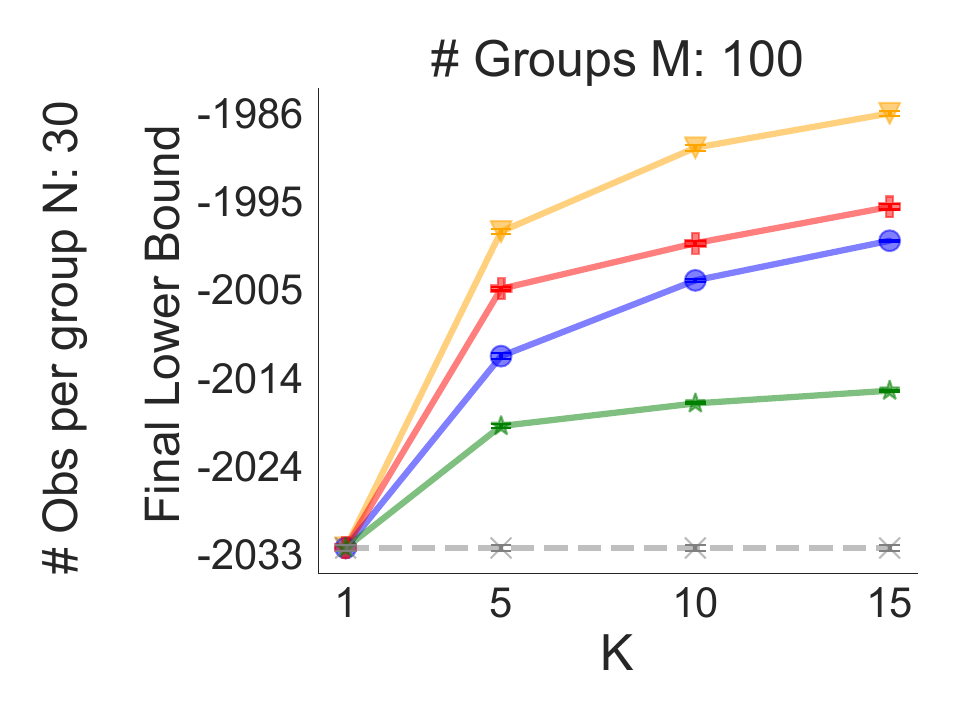}
  
  \vspace{0.5cm}

  \includegraphics[scale=0.4, trim = {0 0 0 1.5cm}, clip]{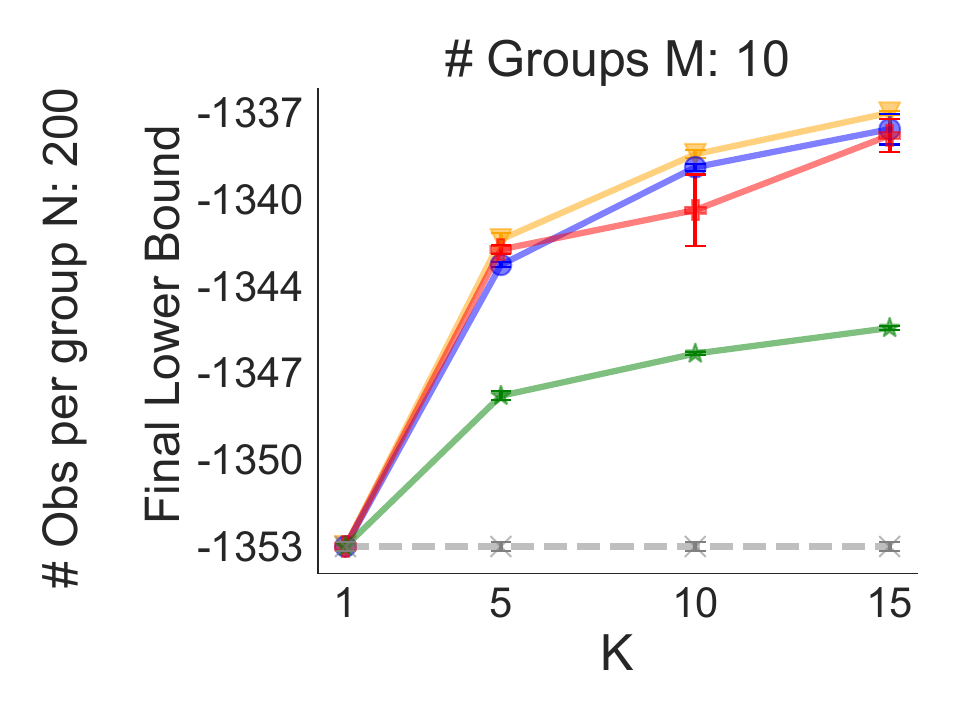}
  \includegraphics[scale=0.4, trim = {3.2cm 0 0 1.5cm}, clip]{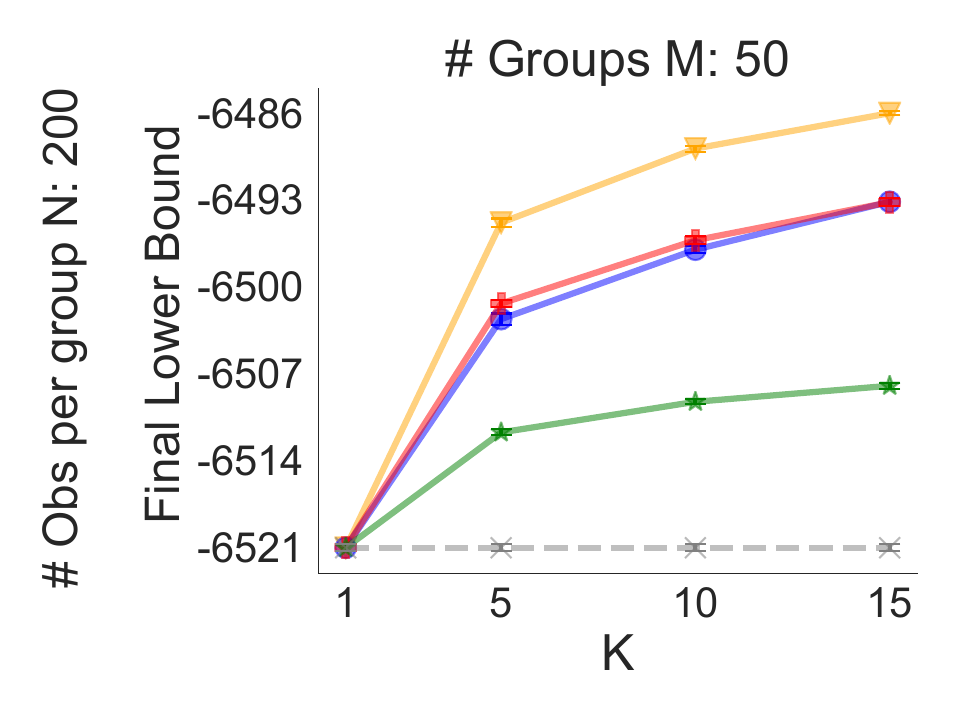}
  \includegraphics[scale=0.4, trim = {3.2cm 0 0 1.5cm}, clip]{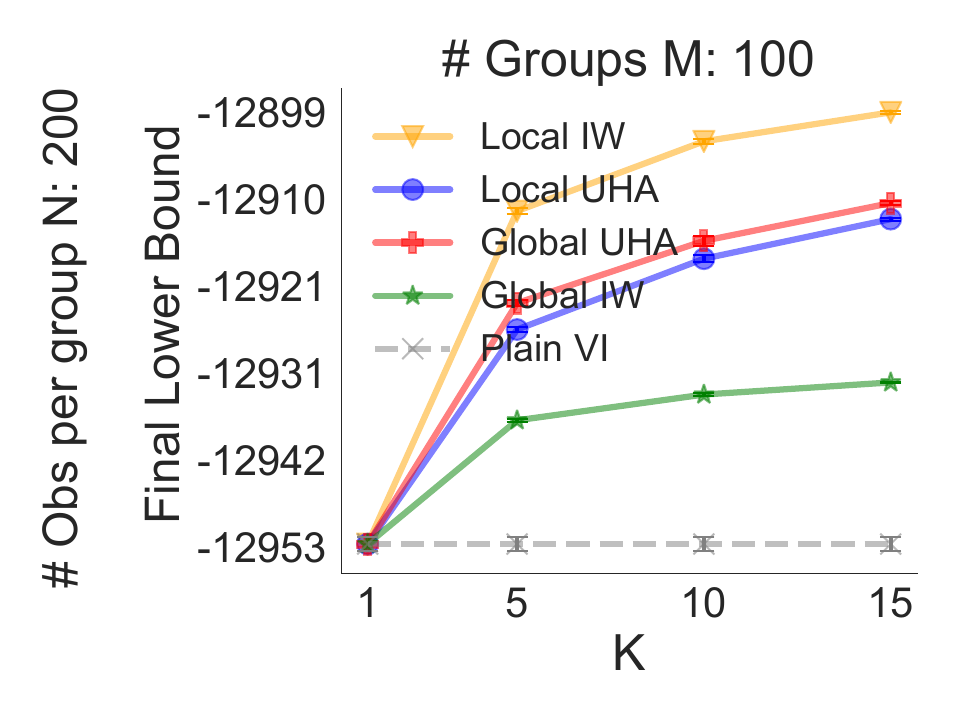}

  \caption{Inference results using locally-enhanced bounds and other baselines on MovieLens datasets of different sizes. The plots show the final lower bound achieved by different methods after training for 50k steps. All methods converge to plain VI for $K=1$.}
  \label{fig:results_mlb_grid}
\end{figure*}

This section presents an empirical evaluation of our new bounding technique. We perform variational inference using locally-enhanced bounds on multiple hierarchical models with real and synthetic datasets. We use a variational distribution $q(\ug, \zg) = q(\ug) \prod_{i=1}^M q(z_i)$, where the approximation for the global variables $q(\ug)$ is set to be a factorized Gaussian, and the local approximations $q(z_i)$ are taken to be independent of $\ug$ and also set to factorized Gaussians\footnote{We parameterize Gaussians using their mean and log-scale.} We test locally-enhanced bounds obtained using importance weighting and uncorrected Hamiltonian annealing for $K\in\{5, 10, 15\}$. We compare against plain VI, which trains the parameters of $q(\ug, \zg)$ by maximizing the $\EVI$ objective from \cref{eq:elbohm} (this corresponds to the ``branch'' approach from \citet{agrawal2021amortized}), and against a direct/global application of importance weighting (objective from \cref{eq:naiveIW}) and uncorrected Hamiltonian annealing. (For the latter two baselines, the tightening methods are applied to all variables, local and global, jointly. See \cref{fig:intuition_IW}.)

We optimize using Adam \cite{adam} with a step-size $\eta=0.001$. For the plain VI baseline and the locally-enhanced bounds we use subsampling with $M'=10$ to estimate gradients at each step using the reparameterization trick \cite{vaes_welling, doublystochastic_titsias, rezende2014stochastic}. We do not use subsampling with the global application of importance weighting and uncorrected Hamiltonian annealing, as the methods do not support it. We initialize all methods to maximizers of the ELBO, and train for 50k steps. All results are reported together with their standard deviation, obtained using five different random seeds.

We clarify that the global importance weighting and global uncorrected Hamiltonian annealing baselines are also trained for 50k steps, using a full-batch approach to compute gradients at each iteration. This results in an optimization process that is significantly more expensive than that of other methods (locally-enhanced bounds and plain VI) which support subsampling.

\subsection{Synthetic data}

\paragraph{Model} We consider the hierarchical model given by
\begin{multline}
p(\mu_z, \psi_z, \psi_y, \zg, \yg) = \mathcal{N}(\mu_z\vert 0, 1) \mathcal{N}(\psi_z\vert 0, 1) \\ \mathcal{N}(\psi_y\vert 0, 1) \prod_{i=1}^M \mathcal{N}(z_i \vert \mu_z, e^{\psi_z}) \prod_{j=1}^{N_i} \mathcal{N}(y_{ij}\vert z_i \, x_{ij}, e^{\psi_{i}}),\label{eq:hmsyn}
\end{multline}
where $x_{ij}$ is external information available for observation $y_{ij}$. In this case $\ug = (\mu_z, \psi_z, \psi_y)$ represents the global variables and $\zg$ the local ones. While the above model defines the local variables $z_i$ to be one dimensional, they can also be defined to have an arbitrary dimension $d_z > 1$, by setting $\mathcal{N}(z_i \vert \mathbf{1}_{d_z} \mu_z, I_{d_z} e^{\psi_z})$ and $\mathcal{N}(y_{ij}\vert z_i^\top x_{ij}, e^{\psi_{i}})$.

\paragraph{Datasets} We generated several datasets by sampling from the hierarchical model above. We consider different number of groups $M \in \{10, 50 ,100\}$, and observations per group $N \in \{10, 30\}$. In all cases, we sample components of $x_{ij} \in \mathbb{R}^{d_z}$ independently from a standard Gaussian.

\paragraph{Results} Results for all the generated datasets are shown in \cref{fig:results_syn_grid}. It can be observed that the use of locally-enhanced bounds yields significant improvements over plain VI, with the performance gap increasing for the larger values of $K$. It can also be observed that the use of locally-enhanced bounds leads to better results than the ones obtained using the global importance weighting baseline. While this baseline is somewhat competitive for the smaller models (left plots in \cref{fig:results_syn_grid}), it is severely outperformed by the use of locally-enhanced bounds in the larger models (right plots in \cref{fig:results_syn_grid}). This is despite the fact that the baseline is significantly more expensive to run, as it does not allow subsampling. On the other hand, we observe that the global uncorrected Hamiltonian baseline performs similarly to the locally-enhanced bound obtained using uncorrected Hamiltonian annealing. However, as mentioned above, this baseline is incompatible with subsampling, thus leading to a significantly more expensive optimization process.

\Cref{fig:results_syn_grid_ll} in \cref{app:ll} shows the test log-likelihood achieved by each method for this model. It can be observed that methods based on uncorrected Hamiltonian annealing achieve the best log-likelihoods (both the global baseline and the locally-enhanced bound), followed by the locally-enhanced bound with importance weighting, and finally by a global application of importance weighting, which performs similarly to plain VI.

\begin{figure}[ht]
  \centering
  \includegraphics[scale=0.55, trim = {2.3cm 0 0 1.5cm}, clip]{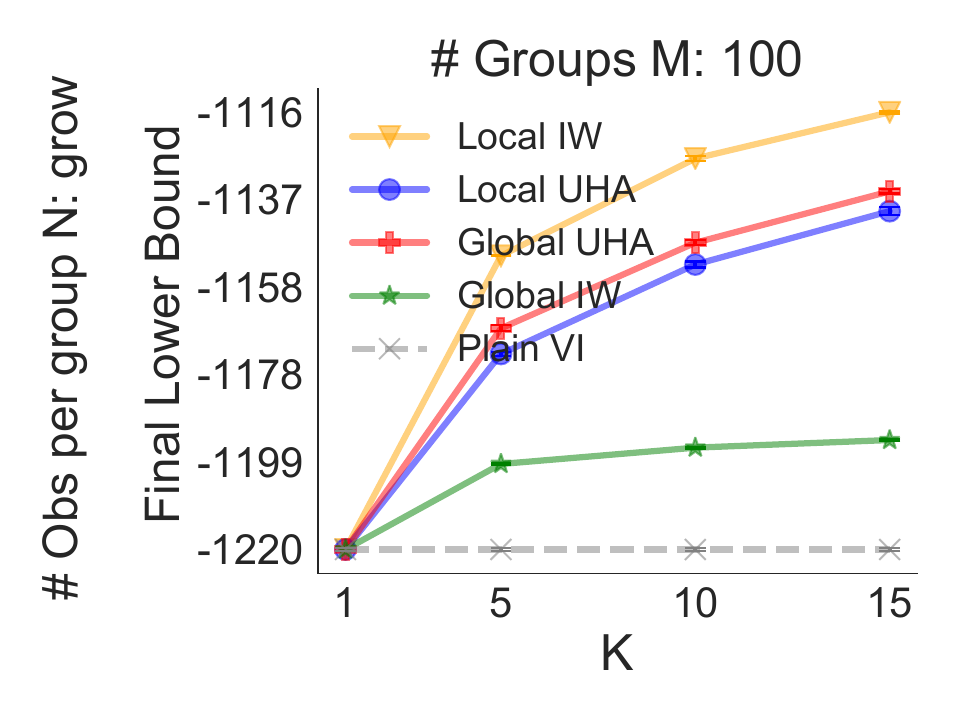}

  \caption{Inference results using locally-enhanced bounds and other baselines on a synthetic dataset generated with $M=100$ groups and a different number of observations per group (see main text). The plot shows the final lower bound achieved by different methods after training for 50k steps. All methods converge to the baseline plain VI for $K=1$.}
  \label{fig:results_syn_grow}
\end{figure}

All results in \cref{fig:results_syn_grid} were obtained for datasets which have the same number of observations for all local groups. To verify the effect that changing this may have, we ran additional simulations using a dataset which contained different number of observations for different groups. Specifically, we considered a dataset composed of $M=100$ groups, out of which 50 have only 2 observations, 30 have 5 observations, and 20 have 30 observations. Results are shown in \cref{fig:results_syn_grow}. Similar conclusions hold.



\subsection{Real data: Movie Lens}

We now show results obtained using data from MovieLens100K \cite{harper2015movielens}. This database contains 100k ratings from several users on $1700$ movies, where each movie comes with a feature vector $x\in\{0, 1\}^{18}$ containing information about its genre. While the original ratings consist of discrete values between $1$ and $5$, we binarize them, assigning $0$ as ``dislike'' to ratings $(1, 2, 3)$, and $1$ as ``like'' to ratings $(4, 5)$.

\paragraph{Model} We consider the hierarchical model given by
\begin{multline}
p(\mu_z, \psi_z, \zg, \yg) = \mathcal{N}(\mu_z\vert 0, I) \mathcal{N}(\psi_z\vert 0, I) \\ \prod_{i=1}^M \mathcal{N}(z_i \vert \mu_z, e^{\psi_z}) \prod_{j=1}^{N} \mathcal{B}(y_{ij}\vert z_i^\top x_{ij}),
\end{multline}
where $x_{ij}\in\{0, 1\}^{18}$ represents the feature vector for the $j$-th movie ranked by the $i$-th user, $\mu_z \in \mathbb{R}^{18}$ and $\psi_z \in \mathbb{R}^{18}$ represent the global variables $\theta$, and $z_i\in\mathbb{R}^{18}$ represents the local variables for group $i$, in this case the $i$-th user.

\paragraph{Datasets} We used data from MovieLens100K to generate several datasets with a varying number of users and ratings per user. Specifically, we consider three different number of users $M\in\{10, 50, 100\}$ and two different number of ratings per user $N\in\{30, 200\}$.



\paragraph{Results} Results are shown in \cref{fig:results_mlb_grid}. It can be observed that the locally-enhanced bound with importance weighting leads to the best results, followed by the locally-enhanced bound with uncorrected Hamiltonian annealing and the global application of uncorrected Hamiltonian annealing (which perform similarly). Finally, similarly to the results observed for the synthetic dataset, a global application of importance weighting and plain VI perform significantly worse.

\section{Discussion and Future Work}

We introduced locally-enhanced bounds, a new type of variational objective obtained by applying tightening methods at a local level for hierarchical models following \cref{eq:hm_simpleobs}. The approach combines the efficiency of plain variational inference and the power of tightening methods while avoiding their drawbacks. An interesting direction for future work involves extending locally-enhanced bounds to more general models that do not follow a two level tree structure. We believe that developing methods able to automatically exploit conditional independences in arbitrary models to build locally-enhanced bounds would be extremely useful in practice.

\section*{Acknowledgements}

This material is based upon work supported in part by the National Science Foundation under Grant No. 1908577. We thank Sam Power for providing the constructions for the augmented distributions presented in \cref{app:div}. 


\bibliography{icml}
\bibliographystyle{icml2022}

\newpage
\appendix
\onecolumn

\section{Proof of \cref{thm:localbound}} \label{sec:proof}

\begin{proof}
We have
\begin{align*}
\log p(\yg) & = \E_{q(\ug)} \log \left( \frac{p(\ug)}{q(\ug)} p(\yg \vert \ug) \frac{q(\ug)}{p(\ug\vert \yg)} \right)\\
& = \E_{q(\ug)} \left[ \log \frac{p(\ug)}{q(\ug)} + \log p(\yg \vert \ug) + \log \frac{q(\ug)}{p(\ug\vert \yg)} \right]\\
& = \E_{q(\ug)} \left[ \log \frac{p(\ug)}{q(\ug)} + \sum_{i=1}^M \log p(y_i \vert \ug)\right] + \mathrm{KL}(q(\ug)\Vert p(\ug \vert \yg)) \\
& = \E_{q(\ug)} \left[ \log \frac{p(\ug)}{q(\ug)} + \sum_{i=1}^M \EG(q(z_i\vert \ug)\Vert p(z_i, y_i\vert \ug)) \right]\\
& \quad + \sum_{i=1}^M \E_{q(\ug)} \left[ \log p(y_i \vert \ug) - \EG(q(z_i\vert \ug) \Vert p(z_i, y_i \vert \ug))\right] + \mathrm{KL}(q(\ug)\Vert p(\ug \vert \yg)).
\end{align*}
In the final equality, note that all terms on the second line are non-negative: the KL-divergence by definition and $\log p(y_i\vert \ug) - \EG(q(z_i\vert \ug) \Vert p(z_i, y_i\vert \ug))$ by assumption.
\end{proof}

\section{Details for AIS and UHA} \label{app:boundsdets}

This section introduces the details for AIS and UHA. We begin with a detailed description AIS, explain how it can be used with HMC transition kernels, and finally move on to UHA, which is built on those ideas.

\paragraph{Annealed Importance Sampling (AIS)} AIS can be seen as an instance of the auxiliary VI framework \cite{agakov2004auxiliary}. Given an initial approximation $q(z)$ and an unnormalized target distribution $p(z)$, AIS proceeds in four steps.
\begin{enumerate}
\item It builds a sequence of unnormalized densities $ \pi^1(z), \hdots, \pi^{K-1}(z)$ that gradually bridge from $q(z)$ to the target $ p(z)$.

\item It defines the forward transitions $T^k(z^{k+1}\vert z^k)$ as an MCMC kernel that leaves the bridging density $\pi^k$ invariant, and the backward transitions $U^k(z^k\vert z^{k+1})$ as the reversal of $T^k$ with respect to $\pi^k$.

\item It uses the transition $T^k$ and $U^k$ to augment the variational and target density. This yields the augmented distributions
\begin{align} 
q(\zgk) & = q(z^1) \prod_{k=1}^{K-1} T^k(z^{k+1}\vert z^k)\\
 p(\zgk) & =  p(z^K) \prod_{k=1}^{K-1} U^k(z^{k}\vert z^{k+1}).
\end{align}

\item It uses the augmented distributions to build the augmented ELBO, a lower bound on the log-normalizing constant of $p(z)$ as 
\begin{equation}
\E_{q(\zgk)} \log \frac{ p(z^K)}{q(z^1)} \prod_{k=1}^{K-1} \frac{U^k(z^{k}\vert z^{k+1})}{T^k(z^{k+1}\vert z^k)}. \label{eq:augELBOAIS_init}
\end{equation}
Then, using that $T^k(z^{k+1}\vert z^k)  \pi^k(z^k) = U^k(z^{k}\vert z^{k+1})  \pi^k(z^{k+1})$, the ratio from \cref{eq:augELBOAIS_init} simplifies to
\begin{equation}
\E_{q(\zgk)} \log \frac{ p(z^K)}{q(z^1)} \prod_{k=1}^{K-1} \frac{ \pi^k(z^k)}{ \pi^{k+1}(z^{k+1})}. \label{eq:augELBOAIS}
\end{equation}
This is the AIS lower bound. Its tightness depends on the specific Markov kernels used, with more powerful kernels leading to tighter bounds.
\end{enumerate}

A particular Markov kernel that is known to work well is given by Hamiltonian Monte Carlo (HMC) \cite{neal2011mcmc, betancourt2017conceptual}. Integrating HMC with AIS is straightforward. It requires extending the initial distribution $q(z)$, the unnormalized target $p(z)$, and the bridging densities $\pi^k(z)$ with a momentum variable $\rho\sim \md(\rho)$. Then, the transitions $T^k(z^{k+1}, \rho^{k+1} \vert z^k, \rho^k)$ and $U^k(z^{k}, \rho^{k} \vert z^{k+1}, \rho^{k+1})$ as an HMC kernel and its reversal, respectively.

It has been observed that Hamiltonian AIS may yield tight lower bounds on the log marginal likelihood \cite{sohl2012hamiltonian, grosse2015sandwiching, wu2016quantitative}. Its main drawback, however, is that, due the use of a correction step in the HMC kernel, the resulting lower bound from \cref{eq:augELBOAIS} is not differentiable, making tuning the method's parameters hard. As we explain next, Uncorrected Hamiltonian Annealing addresses this drawback, building a fully-differentiable lower bound using an AIS-like procedure.

\paragraph{Uncorrected Hamiltonian Annealing (UHA)} UHA can be seen as a differentiable alternative to Hamiltonian AIS. It closely follows its derivation. It extends the variational distribution and target with the momentum variables $\rho\sim\md(\rho)$, augments them using transitions $T^k(z^{k+1}, \rho^{k+1} \vert z^k, \rho^k)$ and $U^k(z^{k}, \rho^{k} \vert z^{k+1}, \rho^{k+1})$, and builds the ELBO using these augmented distributions as
\begin{tequation}
\E_{q(\zgk, \rhogk)} \log \frac{ p(z^K) \md(\rho^K)}{q(z^1) \md(\rho^1)} \prod_{k=1}^{K-1} \frac{U^k(z^{k}, \rho^k\vert z^{k+1}, \rho^{k+1})}{T^k(z^{k+1}, \rho^{k+1}\vert z^k, \rho^k)}. \label{eq:augELBOUHA}
\end{tequation}
The main difference with Hamiltonian AIS comes in the choice for the transition. UHA sets $T^k$ to be an uncorrected HMC kernel targeting the bridging density $ \pi^k(z, \rho)$. This transition consists of two steps, (partially) re-sampling the momentum from a distribution $\mdr(\cdot\vert \rho^k)$ that leaves $\md(\rho)$ invariant, followed by the simulation of Hamiltonian dynamics \textit{without} a correction step. Formally, this can be expressed as
\begin{equation}
\left . \begin{array}{l}
T^k(z^{k+1}, \rho^{k+1} \vert z^k, \rho^k): \vspace{0.1cm} \\
\hspace{0.6cm} 1.\quad \tilde \rho^{k} \sim \mdr(\cdot \vert \rho^k) \vspace{0.1cm} \\
\hspace{0.6cm} 2.\quad (z^{k+1}, \rho^{k+1}) = \mathrm{Dynamics}(z^k, \tilde \rho^{k}).
\end{array} \right . \label{eq:Tuha}
\end{equation}
Similarly, UHA defines the backward transition $U^k$ as the uncorrected reversal of an HMC kernel that leaves $ \pi^k(z, \rho)$ invariant (see \citet{uha} for details).

While closely related to Hamiltonian AIS, the use of uncorrected transition means that $U^k$ is no longer the reversal of $T^k$. Therefore, the simplification for the ratio $U^k / T^k$ used by AIS to go from \cref{eq:augELBOAIS_init} to \cref{eq:augELBOAIS} cannot be used. However, \citet{uha} and \citet{dais} showed that the ratio between these uncorrected transitions yields a simple expression,
\begin{equation}
\frac{U^k(z^{k}, \rho^k\vert z^{k+1}, \rho^{k+1})}{T^k(z^{k+1}, \rho^{k+1}\vert z^k, \rho^k)} = \frac{\md(\rho^k)}{\md(\tilde \rho^k)}, \label{eq:ratiosurprise}
\end{equation}
where $\tilde \rho^k$ is defined in \cref{eq:Tuha}. Then, the bound from \cref{eq:augELBOUHA} can be expressed as \cite{uha, dais}
\begin{equation}
\E_{q(\zgk, \rhogk)} \log \frac{ p(z^K)}{q(z^1)} \prod_{k=1}^{K-1} \frac{\md(\rho^{k+1})}{\md(\tilde \rho^k)}, \label{eq:augELBOUHA_simple}
\end{equation}
which can be easily estimated using samples from the augmented proposal $q(\zgk, \rhogk)$. UHA's main benefit is that, in contrast to Hamiltonian AIS, it yields differentiable lower bounds that admit reparameterization gradients. This simplifies tuning all of the method's parameters, which has been observed to yield large gains in practice \cite{uha, dais}.

\section{Locally-enhanced bounds as divergence minimization} \label{app:div}

Tightening methods can be seen as minimizing a divergence in an augmented space \cite{domke2019divide}. That is, a divergence between an augmented target and variational approximation. This section presents the locally-enhanced bounds obtained with UHA and IW from the perspective of minimizing a divergence in an augmented space, justifying the use of these methods for posterior approximation. Simply put, these constructions follow those by \citet{domke2018importance} and \citet{uha}, but applied at the the local variables' level. These constructions were provided by Sam Power.

\subsection{Locally-enhanced bound with UHA}

This section gives details on the construction of the augmented target and variational approximation that yield the locally-enhanced bound obtained with UHA. This uses a sequence of $K$ densities $\pi_i^1(z), \hdots, \pi_i^K(z)$ bridging between $q(z\vert \ug)$ and $p(z\vert \ug, y_i)$ for each $i=1,\hdots,M$, and follows the construction from \cref{app:boundsdets} but at the local variables' level.

\paragraph{Augmented target} Given the target
\begin{equation} \label{eq:origtarget}
p(\ug, z_{1:M}\vert y) = p(\ug\vert y) \prod_{i=1}^M p(z_i\vert \ug, y_i),
\end{equation}
we apply the UHA augmentation (see \cref{app:boundsdets}) for each local term $p(z_i\vert \ug, y_i)$. That is, we define the $i$-th augmented local term using variables $z_i^{1:K}$ and $\rho_i^{1:K}$ as 
\begin{equation}
p(z_{i}^{1:K}, \rho_i^{1:K}\vert \ug,  y_i) = p(z_i^K\vert \ug, y_i) \md(\rho^K) \prod_{k=1}^{K-1} U^k_i(z_i^k, \rho_i^k \vert z_i^{k+1}, \rho_i^{k+1}, \ug, y_i),
\end{equation}
where the transition $U^k_i$ corresponds to the uncorrected reversal of the HMC kernel targeting the $k$-th ``local" bridging density $\pi_i^k$. Then, the final augmented target is given by
\begin{equation}
p(\ug, z_{1:M}^{1:K}, \rho_{1:M}^{1:K} \vert y) = p(\ug\vert y) \prod_{i=1}^M p(z_{i}^{1:K}, \rho_i^{1:K}\vert \ug,  y_i).
\end{equation}
Note that the marginal over $(\ug, z_{i:M}^K)$ recovers the original target distribution from \cref{eq:origtarget}.

\paragraph{Augmented variational approximation} We define the augmented $i$-th local approximation over variables $z_i^{1:K}$ and $\rho_i^{1:K}$ as
\begin{equation}
q(z_i^{1:K}, \rho_i^{1:K} \vert \ug, y_i) = q(z_i^1\vert \ug, y_i) \md(\rho_i^1) \prod_{k=1}^{K-1} T^k_i(z_i^{k+1}, \rho_i^{k+1} \vert z_i^k, \rho_i^k, \ug, y_i),
\end{equation}
where $T_i^k$ is an uncorrected (underdamped) HMC kernel targeting the $k$-th ``local" bridging density $\pi_i^k(z, \rho\vert \ug, y_i) = \pi_i^k(z\vert \ug, y_i) \md(\rho)$. Then, the final variational approximation is given by
\begin{equation}
q(\ug, z_{1:M}^{1:K}, \rho_{1:M}^{1:K} \vert y) = q(\ug\vert y) \prod_{i=1}^M q(z_i^{1:K}, \rho_i^{1:K} \vert \ug, y_i).
\end{equation}
Then, the marginal over $(\ug, z_{1:M}^K)$ is used as an approximation of the original (un-augmented) target.

\paragraph{Recovering the locally-enhanced bound} Using the augmented distributions defined above we get the following decomposition of the marginal likelihood
\begin{equation}
\log p(y) = \EVI\left(q(\ug, z_{1:M}^{1:K}, \rho_{1:M}^{1:K} \vert y) \Vert p(\ug, z_{1:M}^{1:K}, \rho_{1:M}^{1:K} \vert y)\right) + \mathrm{KL}\left(q(\ug, z_{1:M}^{1:K}, \rho_{1:M}^{1:K} \vert y) \Vert p(\ug, z_{1:M}^{1:K}, \rho_{1:M}^{1:K} \vert y)\right),
\end{equation}
where the first term in the RHS is exactly the locally-enhanced bound obtained with UHA. Since $\log p(y)$ is constant, maximizing the locally-enhanced bound is equivalent to minimizing the KL-divergence between the augmented distributions. Finally, the chain-rule of the KL-divergence \cite{cover1999elements} justifies the use of the marginal $q(\ug, z_{1:M}^{K} \vert y)$ as the approximation to the original target distribution.

\subsection{Locally-enhanced bound with IW} The derivation follows that of \citet{domke2018importance}, but applied at the local level (as done above for UHA). We refer the reader to the work by \citet{domke2018importance} for details regarding the construction.


\section{Test log-likelihoods for synthetic model} \label{app:ll}

\begin{figure*}[ht]
  \centering
  \includegraphics[scale=0.4, trim = {0 2.4cm 0 0}, clip]{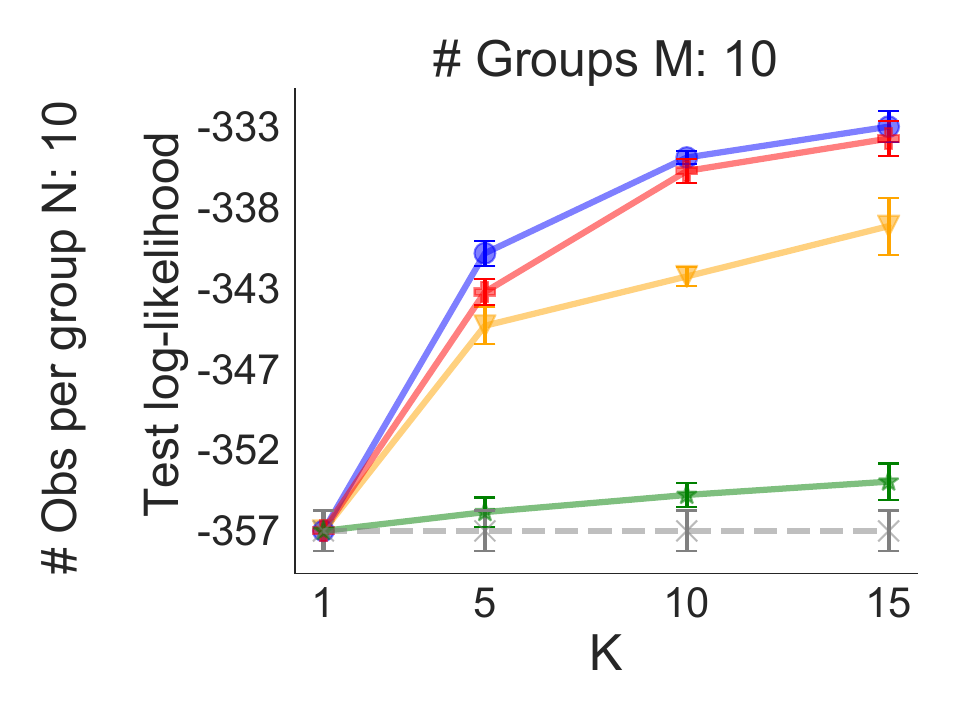}
  \includegraphics[scale=0.4, trim = {3.2cm 2.4cm 0 0}, clip]{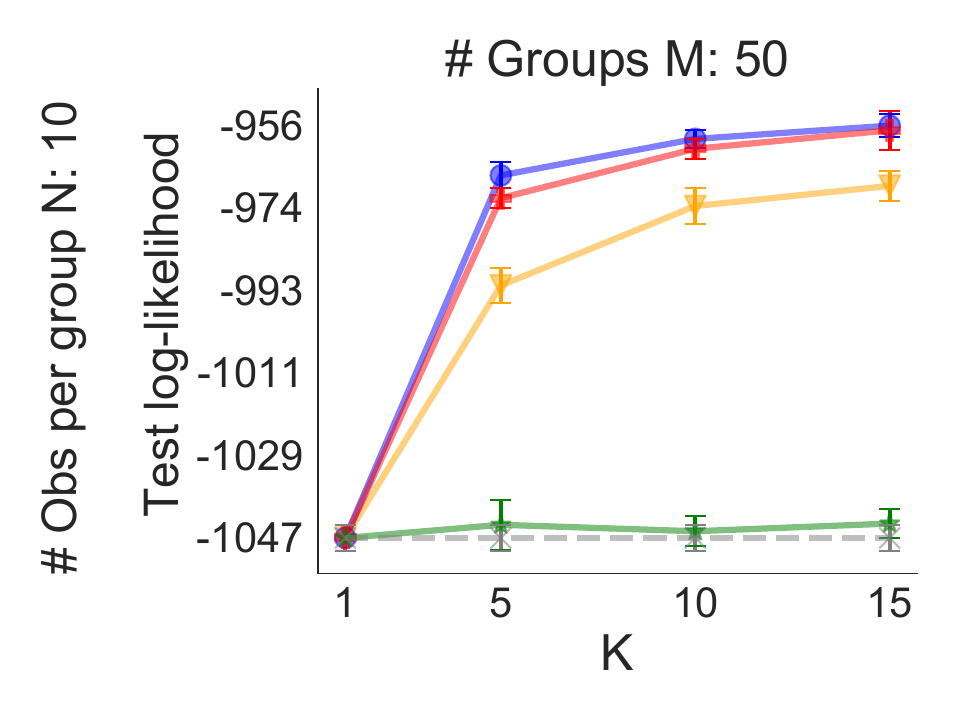}
  \includegraphics[scale=0.4, trim = {3.2cm 2.4cm 0 0}, clip]{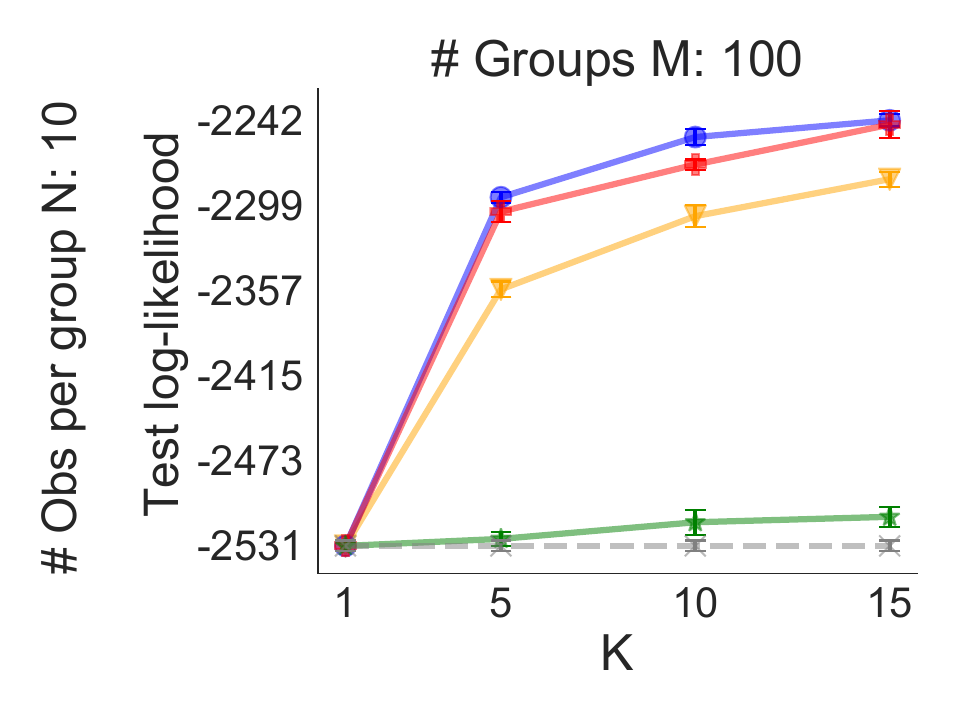}

  \vspace{0.5cm}

  \includegraphics[scale=0.4, trim = {0 0 0 1.5cm}, clip]{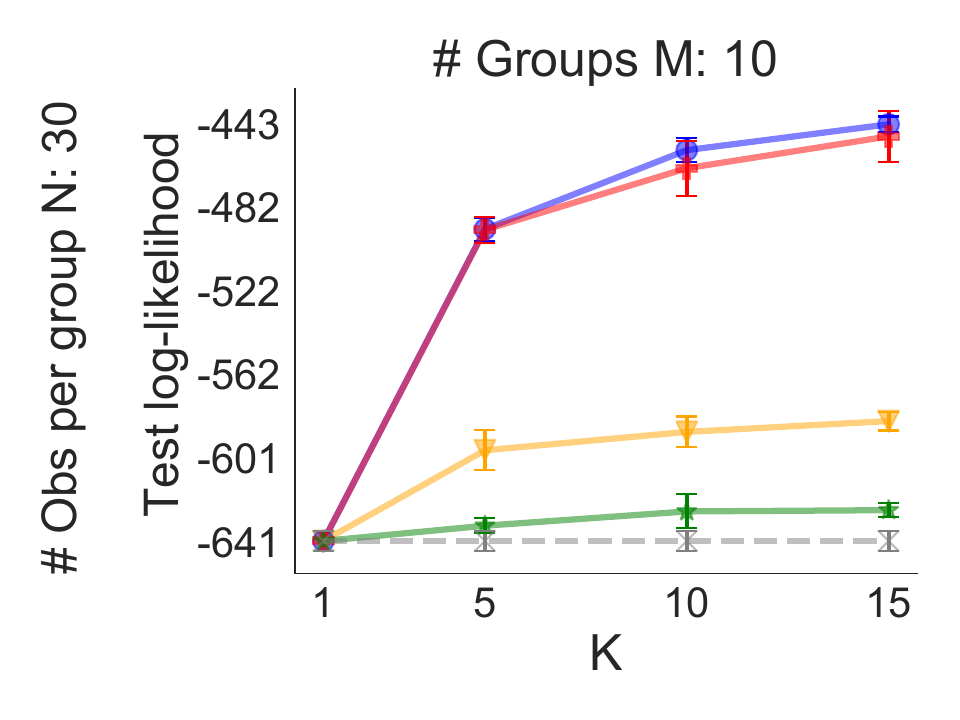}
  \includegraphics[scale=0.4, trim = {3.2cm 0 0 1.5cm}, clip]{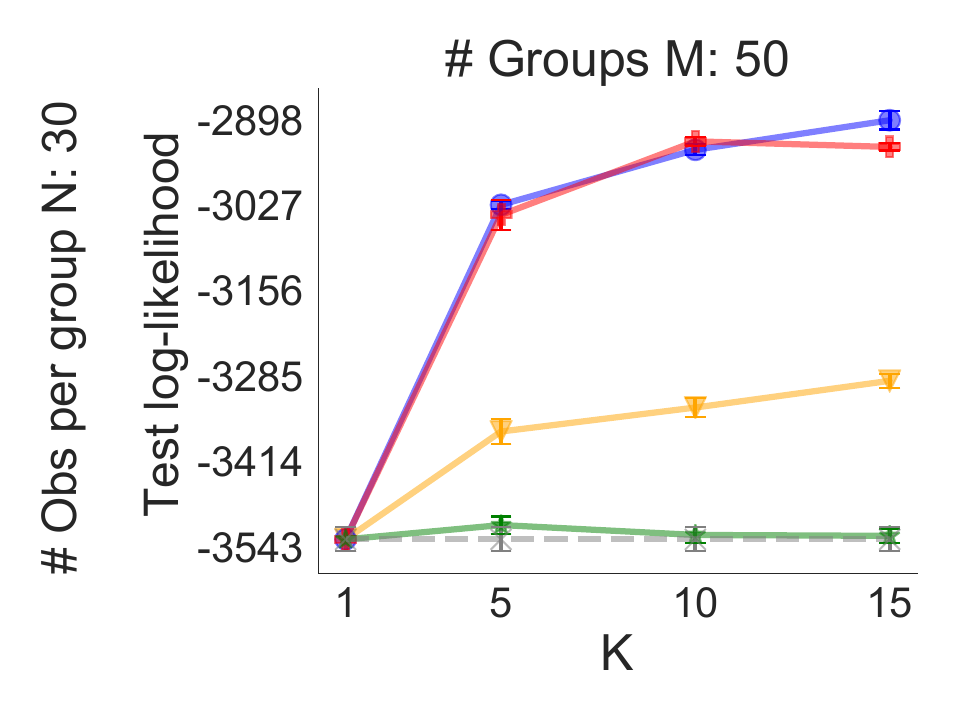}
  \includegraphics[scale=0.4, trim = {3.2cm 0 0 1.5cm}, clip]{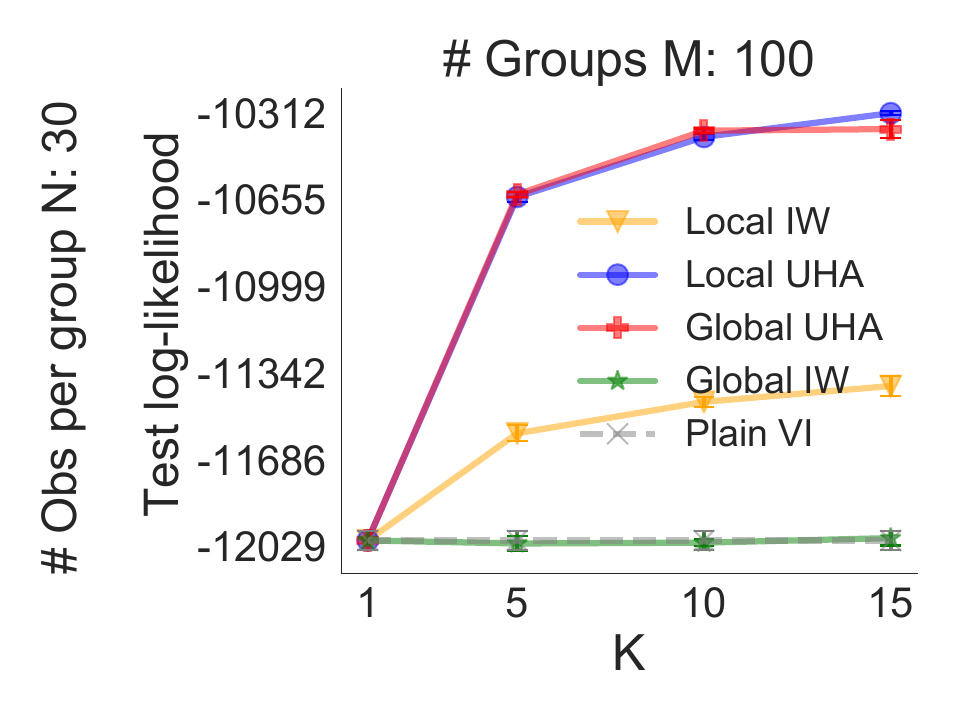}

  \caption{Test log-likelihoods achieved by each method on synthetic datasets of different sizes. The plots show the test log-likelihoods achieved by different methods after training for 50k steps. All methods converge to plain VI for $K=1$. The dimensionality of the local variables $z_i$ is taken to be $d_z=5$ for the datasets with $N=10$ observations per group, and $d_z=20$ for the datasets with $N=30$ observations.}
  \label{fig:results_syn_grid_ll}
\end{figure*}

\Cref{fig:results_syn_grid_ll} presents log-likelihoods obtained by each method on a held-out test set. 

\end{document}